**ORIGINAL PAPER**

# A framework for information extraction from tables in biomedical literature

Nikola Milosevic[1] · Cassie Gregson[2] · Robert Hernandez[2] · Goran Nenadic[1]

## Abstract

The scientific literature is growing exponentially, and professionals are no more able to cope with the current amount of publications. Text mining provided in the past methods to retrieve and extract information from text; however, most of these approaches ignored tables and figures. The research done in mining table data still does not have an integrated approach for mining that would consider all complexities and challenges of a table. Our research is examining the methods for extracting numerical (number of patients, age, gender distribution) and textual (adverse reactions) information from tables in the clinical literature. We present a requirement analysis template and an integral methodology for information extraction from tables in clinical domain that contains 7 steps: (1) table detection, (2) functional processing, (3) structural processing, (4) semantic tagging, (5) pragmatic processing, (6) cell selection and (7) syntactic processing and extraction. Our approach performed with the $F$-measure ranged between 82 and 92%, depending on the variable, task and its complexity.

## 1 Introduction

The literature in the biomedical domain is growing exponentially. Currently, there are over 26 million articles indexed in MEDLINE [35]. In 2015, on average, around 2200 new papers were published each day [43].

Researchers and professionals are no more able to cope with this amount of literature. Fields of text mining and natural language processing provide tools and methodologies that can help with retrieving relevant information. However, most of the current approaches are limited to the textual body of articles, usually ignoring figures, tables and other semi-structured presentation formats of information. Also, text mining methods applied to tables often perform poorly. Authors of the scientific literature use tables to present detailed information about the settings and the results of their experiments. Tables are used also for other purposes, where authors need to present a relatively large amount of multidimensional information in a compact manner [42]. Tables contain essential information needed for reproducibility of research and comparison to other studies.

Tables can be complex to understand, and even human readers struggle to understand the information presented in them [50,51]. Wright [51] pointed out the challenge of reading implicitly stated information from the table, where the reader is required to perform a mental operation (usually some kind of calculation) in order to obtain all the necessary information. The main challenges for understanding the tables are:

- *Variety of structural layouts and visual relationships* The structure of the table is determined by the structure and the relationships of its cells. One cell can span over several cells both vertically or horizontally, and combinations of spanning cells can create a vast number of structural variation. Also, some emphasis features of text and table lines can affect the way tables' structure is understood. For example, horizontal lines or bold text may emphasize multiple headers of the table. The structure of the table visually defines the relationships between cells. Relationships between information in the table are visual, while, on the other hand, relationships between words in the text are linear. Visual relationships in tables make it difficult to

✉ Nikola Milosevic
  nikola.milosevic@manchester.ac.uk

1  School of Computer Science, University of Manchester, Manchester M13 9PL, UK

2  AstraZeneca plc, Cambridge, UK

computationally find the related cells and extract information from them.

- *Representation for visualisation* Most of the representation formats for tables, such as markup languages in which tables can be described, are designed for visualisation. Therefore, it is challenging to automatically process tables.
- *Variety of value presentation patterns* Values in cells can be presented using different syntactic representation patterns. For example, the mean and standard deviation can be represented using a form with $\pm$ sign (i.e. $16 \pm 2$) or standard deviation can be represented in the bracket (i.e. 16 (2)). Extraction of numerical values requires knowledge of possible presentation patterns.
- *Dense content* The content of the cells can be either numerical or textual. However, textual content is usually dense, containing ambiguous short chunks of text with the use of acronyms and abbreviations. This is especially true in biomedical publications. In order to understand tables, the text needs to be disambiguated and abbreviations and acronyms need to be expanded.

Presented challenges, in addition to natural language processing challenges, make it hard to understand the structure and the information that the table introduces. Information extraction from tables requires multilayered analysis that will include functional, structural, pragmatic, syntactic and semantic analysis.

Our research is focusing on the task of extracting numerical and textual information from tables. In this paper, we present a framework for information extraction from tables in biomedical documents. In this framework, we modelled variable types in tables, proposed a recipe for creating successful table information extraction systems, with the prescribed necessary knowledge needed about the variable that should be extracted. The methodology and framework are validated on the task of extracting baseline characteristics of the patients per each clinical arm (number of patients, average age, gender distribution), adverse events they encountered and extraction of drug–drug interactions from drug product labels. For example, looking at the table from Table 1, our method should extract that there were 42 female participants in the placebo group and 34 female participants in Mannitol group. Also, the method should be able to extract that there were 52.5% of female participants in the placebo group and 44.7% of female participants in Mannitol group. We focus the validation of the method on baseline characteristic data from clinical trials; however, the approach can be used in other areas (such as mining drug labels). The framework defines classes of variables, prerequisite knowledge about them, the steps and analysis layers for extracting information from tables. We also compare a machine learning approach to a rule-based approach to identify cells with information of interest and evaluate how and where machine learning can help efficient information extraction from tables.

## 2 Background

Tables are viewed and manipulated for several different purposes (i.e. creating/editing, reading, mining). This has led to the specific models that give an insight about important tables characteristics from these viewpoints. A table is a representation of organisation (layout), structure and content of tables.

Tables can be considered at three levels of description: abstract, physical and logical [25].

- *The abstract level* encapsulates the communicative intent of the author (i.e. relationships between the data) [45].
- *The physical level* consists of pixels, lines, and text located in documents or other display devices. They are referred as layout structures of tables [15,18].
- *The logical level* describes the arrangement and content of the table elements. Tables at the physical level are usually described and created on a logical level using some descriptive language, such as HTML, XML or LaTeX [15,33,45].

Hurst in his work presented a model of tables containing 5 components: graphical, physical, functional, structural and semantic [19]. With his model, Hurst also defines a workflow for information extraction from tables. Previous work done in table mining, including Hurst's, can be classified into three main tasks: (1) table detection (usually examining transformation from a graphical and physical presentation

**Table 1** Example of a baseline characteristic table reporting number of participants (PMC2147028)

| Variable | Placebo $N = 80$ | Mauuitol $N = 76$ | $P$ value |
|---|---|---|---|
| Female | 42 (52.5%) | 34 (44.7%) | 0.33 |
| Fever | 79 (98.8) | 76 (100%) | 0.33 |
| Convulsions | 79 (98.8%) | 75 (98.7%) | 0.97 |
| Duration of coma | 7.0 (IQR3.5–12.0) | 6.0 (5.0–12.0) | 0.79 |
| Blantyre coma score 1/5 | 13 (16.2%) | 10 (13.2%) | 0.59 |

Baseline clinical characteristic of the 156 patients with cerebral malaria in both treatment arms on admission and identification of tables), (2) functional and structural analysis, (3) table understanding (examining a semantic component of the model).

Detecting tables is a task that is relevant for certain types of documents, such as PDF or plain text (ASCII) documents. Detecting tables in PDF documents usually relies on optical character recognition [21] in combination with machine learning or heuristics approach [8,53]. Detection of tables in plain text documents relied on the spatial arrangement of text as features, machine learning algorithms, such as decision trees [34], or heuristics. In XML-like documents, it can be trivial, unless tables are used for document formatting. For example, in HTML documents, during the last decade of the twentieth century and the first decade of twenty-first century, it was common to use tables for arranging page elements, which was later replaced by division elements (div tags) due to the focus on the responsive design of web pages. In HTML documents that are formatted using tables, it is necessary to discriminate which tables are genuinely used for presenting data and which ones are used for layout. This can be done using spatial characteristics of the table (spanning cells, number of columns and rows, etc.), content formatting (bold, italic) and content as features for machine learning algorithms such as support vector machines and decision trees [41,47].

In order to understand tables, it is necessary to understand the functions of the areas in the table. Recognizing functional areas (headers and data areas) has been done mainly using machine learning methods like decision trees [6] or sequence modelling approaches such as conditional random fields [48]. The consistency of cells over the column can be used as an indicator for the detection of headers [20]. Silva [39] argued that there is no single algorithm that is good enough to successfully determine the functional areas of tables and proposed to use multiple machine learning and heuristics algorithms either in sequence or in parallel.

The Biotext Search engine is the example of specifically crafted search engine based on Lucene that is able to index and assign different weights to the textual chunks in cells and caption of the tables [17]. TableSeer is a table search engine that incorporates TableRank algorithm that is taking into the account the importance of the table in the document, impact factor of the journal in which table is published and author information in order to create a ranking of the tables for a certain search term [24]. Chen et al. [7] modelled information extracted from tables as key values pair, but also noted that value can have multiple attributes that define it (multi-dimensional tables), so they proposed to merge multiple attribute-value pairs in a sense that attribute in one pair may act as a value for the other. Another model of extraction template was to present it as $\langle p, s, o \rangle$ triple, where $p$ is relation or predicate, $s$ is a subject and $o$ is an object [9]. Assigning a concept to terms in tables is possible by using overlapping triplets, extracted from table columns and clustering them [10]. Named entity recognition can be developed using available knowledge sources such as DBPedia, Free-Base, WordNet, Yago [32] or thesauri and ontologies [46]. Wei et al. [48] presented question answering approach for tables based on machine learning labelling of functional areas that were transformed into cell documents. These cell documents were indexed, and information retrieval methodology was used in order to select the most relevant cell for the given query. Wong et al. [49] presented an approach to extract gene mutation names from tables in biomedical papers by using a database of known gene mutation names (MMR). In case content of multiple cells in the certain column was matched with the entries in the database, they assumed that whole column present gene mutation names. The unknown gene mutation was extracted and added to the database.

Despite the amount of the research presented in the field of table mining, there is still no integrated approach to information extraction that would be able to handle all types of tables in a certain domain. Extraction templates are also created with simple tables in mind, and many of functional analysis research focus on column header detection only. Most of the current approaches are utilizing information retrieval and named entity recognition methodologies, but they rarely focused on extracting numerical data from tables.

## 3 Methodology

### 3.1 Extraction template

The goal of the information extraction task is to extract certain variable and store extracted information in a defined template. For tables, we propose the following extraction template:

$$(VariableName, VariableSubCategory,$$
$$ValueComponent, Context, Value, Unit)$$

- *VariableName* is the name of the variable that should be extracted. It can be linked with a certain ontology (e.g. Ontology of Clinical Research (OCRe) [40] or UMLS).
- *VariableSubCategory* is used only for variables when there are multiple subcategories that have values (e.g. ethnicity and number of participant presented as a number of White, Asian, Hispanic and Black people).
- *ValueComponent* parameter presents the name of the value component of the extracted variable's value, obtained by analysing its presentation pattern. For example it may be *Value* if the cell presents a single value, *Range:Min* if the extracted value is minimum in the range, *Range:Max* for the maximum in the range,

**Table 2** Categories of information that need to be described in order to specify table information extraction task

| Descriptor name | Description | Example |
| --- | --- | --- |
| Semantic identifier | Describes the way of mapping between the information and certain knowledge source | age as UMLS:C0001779 |
| Table's pragmatic type | Pragmatic type of a table in which information is likely to appear | Pragmatic types can be for example tables with Baseline Characteristics, Adverse events, Inclusion/Exclusion, etc. |
| Cues | | |
| Lexical cues | Set of lexical cues and patterns that determine whether the value is present in certain cells | Lexical cue for number of patients can be "$n = \%d$", "number of patients" in stub or number in data cell |
| Functional cue | Description of functional regions in the table where information may appear | Number of patients may be in caption, header or data cell |
| Semantic cues | Set of semantic cues such as semantic types and higher level concept names | List of semantic types indicates the presence of the value (i.e. The Sign or Symptom UMLS semantic type may indicate an adverse event in table) |
| Value type/pattern (syntactic cue) | Description of the value type and its pattern with the way to extract it | Whether the value is single number, range, percentage, etc. |
| Unit of measure | Description of the unit of measure and recognition cues. Definition of the default unit for the information class | Default is gram (g), but kilogram (kg) and milligram (mg) may appear |

*Percentage* for values presenting percentage, *Mean* for mean values, and *SD* for standard deviation. In the case when a cell presents a range, two rows in the template should be extracted, one for the minimum and one for the maximum.

– The *Context* is the parameter that describes the value's context. It can be, for example, a clinical trial arm for tables presenting cumulative baseline characteristics of patients, or a patient identifier for tables presenting baseline characteristics for each patient separately.

– The *Value* is the extracted value for the given variable from the table.

– The *Unit* parameter is only applicable for numeric variables, where it is used to specify the unit of measure in which the value is expressed. For example, body mass can be presented using a singular unit (gram), multiples (kilogram) or sub-multiples (milligram) [44]. Each variable should have defined a default unit (if it exists, usually it is a singular unit) and that unit is used if it is not otherwise specified in the table.

Additionally, the template should retain a bond to the article and table from which the information is extracted.

### 3.2 Information extraction task specification

Before implementing information extraction, it is necessary to describe the task and specify the variable that should be extracted. We present a template that describes the task and variable for extraction. The template contains seven description categories that need to be defined. The categories are presented in Table 2.

First, it is necessary to define what we want to extract. We can define that using semantic identifier and map the variable name to some knowledge source. It is usual that certain types of information are grouped together in a certain type of tables. Pragmatic type of table define what is the table used for and what is usual information that is stored in it. Defining pragmatic type of the table narrows the scope of the task and then tries to extract information only from relevant tables. Further, variables need to be located, for which are used lexical, functional and semantic cues. It is necessary to define the meaning of the values that are parts of the numeric expression. For example, the value may present the mean or median with standard deviation using the same syntactic pattern. It is possible to identify whether the value is mean or median by checking the content of the stub cell in the same row. However, the majority of tables present mean without explicitly stating it. Thus, default value helps to assume that the meaning of the value is "mean value" in this case. For some of the description categories, it is useful to define a default value. Often tables present values but without its unit. A unit is one of the values for which it is useful to define a default value. However, it is also necessary to define a procedure for extracting and checking the unit.

### 3.3 Information groups

Our model of table information contains five information groups whose extraction methodology slightly differ

**Fig. 1** Diagram of information types having different patterns and extraction methodologies

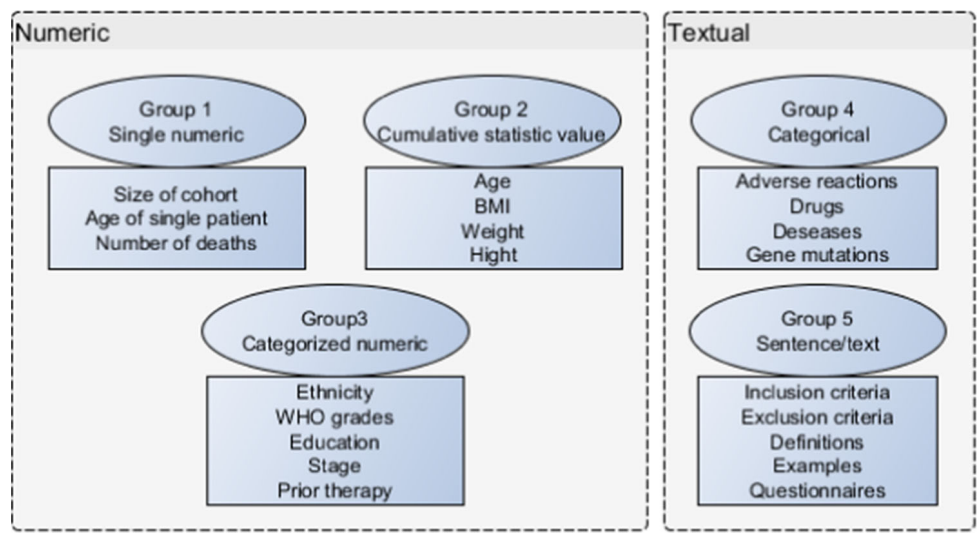

(see Fig. 1). They are grouped into two high-level types—(1) numeric and (2) textual information groups. There are three numerical groups—(1) single numerical value, (2) aggregated statistical value and (3) categorized numeric values. There are two textual groups—(4) categorical and (5) free text information classes.

### 3.3.1 Numeric information groups

The numerical variable types contain three subtypes:

**Group 1—Single numeric** The first group represents the values represented as a single numerical value (e.g. *15, 24.3*). In demographic tables in the clinical trial literature, this may be the size of cohort if we examine tables presenting aggregated data about an entire cohort or age, BMI, weight or the height of a single patient if the data are presented per participant. Individual measurement results are often presented using single numerical subtype.

**Group 2—Aggregated statistical values** Demographic data in tables are often presented cumulatively, for the whole cohort or for the groups participating in trial arms. In such cases, values are usually presented as the mean value with optional standard deviation or range (e.g. $15.3 \pm 2.1$, $24\,(14 - 35)$, $16 \pm 2\,(14 - 17)$). Examples of information from this group are BMI, weight, height and/or age of patients in aggregated demographic tables.

**Group 3—Categorized numeric values** Values in this group have multiple subcategories and are presented as numbers, means, ranges or percentages per subcategory. Examples of such values are ethnicity (e.g. *number of White, Asian, Black, Hispanic, etc.*) or the number or percentage of patients with a certain stage of disease, adverse reactions, etc. For this subtype, it is necessary to define possible categories and the mapping between the category names and cues identifying them in tables. Numerical values categorized by two categories are a special case since they can be presented in a single cell (e.g. 27/28). An example of such information is the gender of participants in a clinical trial. In some cases, values are presented in multiple rows —typical for this group— while in other cases binary categorized values are presented using special presentation pattern (i.e. explicit pattern, such as *"male/female—22/14"* or implicit, such as *"female (%)—14 (39%)"*). Category cues are usually in navigational areas (headers or stubs) but in some cases can be in data cells (e.g. *14 M, 18 F*).

### 3.3.2 Textual variables

Textual information can be grouped into the two groups:

**Group 4—Categorical values** Categorical values are controlled words or short phrases, such as names of diseases, adverse reactions, drugs, institutions, etc.

**Group 5—Free text** The last group presents free text information. Examples of such information are inclusion and exclusion criteria, the definition of terms or scales and examples of questions asked in a questionnaire. They are longer phrases, sentences or even paragraphs of texts stored in tables. Following extraction, they can be further mined using standard text mining techniques. However, free text variables are outside the scope of this paper.

## 3.4 Information extraction methodology

The approach consists of seven steps: (1) table detection, (2) functional processing, (3) structural processing, (4) semantic tagging, (5) pragmatic processing, (6) cell selection, (7) syntactic processing and extraction. As a dataset to test our method, we used clinical documents stored as open access in PubMedCentral.[1] Overview of the methodology is presented in Fig. 2.

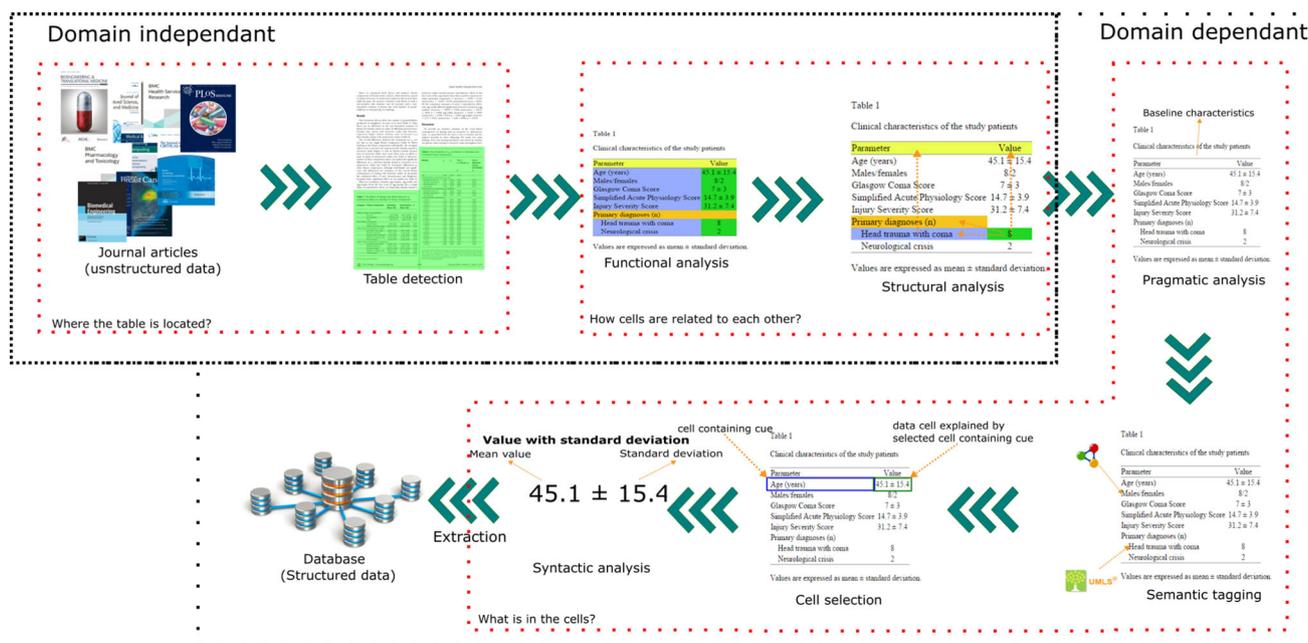

**Fig. 2** Overview of the methodology steps

### 3.4.1 Table detection

In table mining, it is necessary to identify table mentioned in documents. In some types of documents, this task might be trivial. For example, in many XML formats, it is possible to identify tables by extracting the content of a specific XML tag. However, in some types of XML documents and in HTML it is a much harder task. Research on the identification of tables from HTML was described in [12,41,47] and is outside the scope of this paper. Our methodology is developed for documents in PMC database [37], where tables have been identified by locating appropriate *table* tags.

### 3.4.2 Functional processing

The aim of functional processing is to detect the basic roles of cells. The cell can be column header, row header (stub), super-row (row or part of row header that categorize additionally row header) or a data cell. In order to detect functional roles of the cells, we used a set of heuristics about cell positions, its neighbours, content type, surrounding XML tags and XML attributes (such as span). Functional analysis method was explained in more details in [30].

### 3.4.3 Structural processing

During the structural processing, the relationships between cells are recognized, which include relationships to the navigational cells such as headers, stubs, and super-rows. We used a set of heuristics about cell's function, structure, content, position and table's structure to disentangle table's structure and inter-cell relationships. Information about cell's content, position, function, and relationships are stored in a database.

For structural processing, we applied a previously described heuristic approach [30].

### 3.4.4 Semantic tagging

Once data are stored in the database, we enrich the data by annotating the cell's content by using named entity recognizers and vocabularies, such as UMLS [5]. We used MetaMap [3] to annotate content with concept ids and semantic types from UMLS. Metamap and UMLS provide annotations for a wide variety of concepts and semantic types in the biomedical domain because UMLS encapsulates almost 200 biomedical controlled vocabularies and classification systems, including ICD-10, MeSH, SNOMED-CT and Gene Ontology.[2] For further annotation purposes, we developed a dictionary-

---

based concept tagging method that is able to annotate text using UMLS (MetaMap), WordNet [28], DBPedia [4] and vocabularies represented in Simple Knowledge Organization System model [27,29]. In the case of clinical trial publications, the most useful annotation was provided by UMLS; however, in other applications, other semantic knowledge sources may be more useful.

### 3.4.5 Pragmatic processing

Pragmatics is the study of how context and the way information is communicated contributes to meaning [23]. In the case of tables in the literature, we consider pragmatics to analyse the author's intentions regarding the context and the purpose of the table.

Usually, authors intentionally group certain information, such as demographic information, adverse events or inclusion and exclusion criteria. The main purpose of pragmatic analysis is to identify a target table where the variable we are extracting and reducing the number of false positives. In our methodology, we design table pragmatic analysis as a table level annotation task.

There may be many different reasons which information authors decide to group together and why. Therefore, there may be a very large amount of pragmatic table types. However, we have adopted a strategy where the system designer only identifies table types of interest. For example, if one needs to design the system extracting baseline trial characteristics and adverse events, he can create a system identifying three types of tables: baseline characteristics, adverse events and other.

Pragmatic analysis of the tables can be performed using a rule-based and machine learning approach, depending on the structure of the analysed documents. For example, in drug labels presented in the DailyMed database,[3] it is possible to develop rules that will select only tables in a certain section (e.g. *drug interactions, adverse reactions, dosage and administration,* etc.), using section identifier (for identifying sections DailyMed uses LOINC—Logical Observation Identifiers Names and Codes). The drug labels are well structured into topic-related sections where relevant tables can be found. However, in different scientific publications tables presenting the same variable group (e.g. baseline characteristics or adverse reactions) can be in different sections. Therefore, it is not possible to select relevant tables based on rules.

For documents where it is challenging to develop a rule-based approach for pragmatic analysis, we propose a machine learning classification method that analyses captions and the variables presented in a table, with an aim to determine the purpose of a given table and the types of information stored in it. Since the proposed methodology utilises supervised machine learning, firstly, it is necessary to define the classes of tables and manually annotate a set of tables to be used as a training set. The classes of tables should reflect variable groups that are commonly presented together (e.g. baseline characteristic variables, such as a number of patients, their age, gender, weight, height, body mass index, are commonly presented in one table). Defining pragmatic table classes may take into consideration potential future tasks (e.g. information extraction or retrieval). Once the pragmatic classes are defined, it is necessary to annotate a set of tables that are later used for training machine learning method. For training machine learning algorithms on the described features, we used Weka toolkit [14] with default parameters, while utilizing TF-IDF transformation and snowball stemming.

In order to test and evaluate how different parts of the table contribute to pragmatic classification, we designed a case study of clinical trial articles. Since we consider mainly extraction of baseline characteristics (patient number, age, gender etc.) and adverse reactions, table classes reflect these requirements. Possible tables classes in our experiment were *"baseline characteristics"*, *"adverse events"*, *"inclusion/exclusion criteria"* and *"other"*. As features, we used words from the caption, column and row headings, sentences referring to a table, the number of rows and number of columns that the table has. The features are extracted from the XML structure of the document and table. This was performed using TableDisentangler methodology described in [30].

### 3.4.6 Cell selection

Once the knowledge about the variable or recipe for the information class is provided by the user, the framework method can extract the defined variables and their values.

In the cell selection step of the methodology, cells are analysed and the information is extracted. Firstly, cells are retrieved. The second step performs the analysis whether the information that we are looking for is contained in a particular cell. The selection and analysis step can be performed either by using heuristics or machine learning. At the end, our method performs the analysis over the cell's value, assigns semantics to the value, extracts it and fills the extraction template. The diagram of our methodology is presented in Fig. 3.

Cell retrieval is dependant on the cell selection strategy. In the case of the machine learning approach, all table's cells are retrieved and whole cell analysis is performed by the machine learning algorithm. In the case of the rule-based strategy, it is possible to include parts of the analysis in the cell retrieval step. This can be done by retrieving the cells that contain themselves or in their navigational areas certain lexical cue. For example, if we want to extract the age of patients in the

---

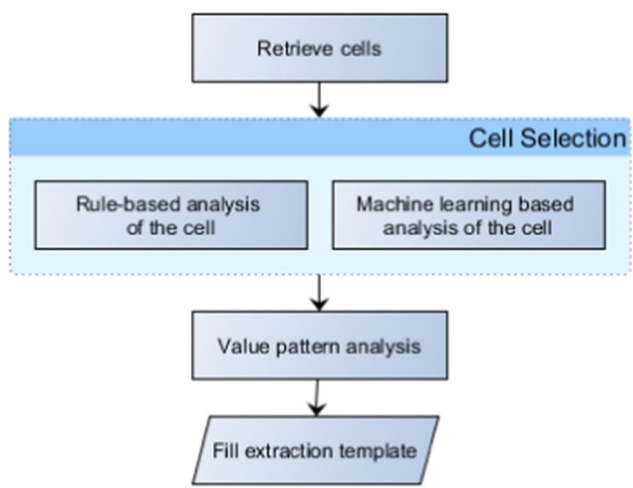

Fig. 3 Diagram of cell selection, syntactic processing and extraction steps of the methodology

clinical trials, we will select cells containing age in cell's navigational path (headers, stubs, super-row).

Two approaches for cell selection are possible:

– *Heuristic-based approach* Heuristic-based approach is already started by selecting only cells that contain certain lexical cue in its context in the previous step. Further, we analyse the content of the cell and related navigational cells. The method is looking for lexical cues that indicate the existence of the information in the selected cell. Lexical cues that indicate the existence of certain information in the analysed cell are defined in the lexical white list. On the other hand, some words can modify the semantics of the cell, even if it contains the searched cue. In this cases, we need to discard the selected cell. For example, if we are looking for BMI, a cell that contains as content "BMI change" is not of interest. These cues modify the meaning of the cell, so the information in them should not be extracted. Such cues are defined in the blacklist. The method is also able to analyse whether the value presentation pattern matches the usual pattern for presenting that kind of information by using regular expressions. The heuristics need to be crafted manually based on the previously crafted information description and improved by using insights from the data. The improvement process is performed by selecting a certain number of random tables as a training set, running the heuristics on them and iteratively improving them until the results are satisfactory.
– *Machine learning-based approach* Machine learning cell analysis classifies cells into the ones containing values of variables for extraction and the ones not containing values of interest. In this approach, it is necessary to select a certain number of random tables and annotate

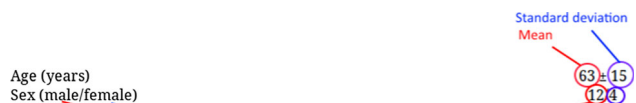

Fig. 4 Syntactic analysis infers the implicit meaning from the value presentation pattern (upper row) or link to the explicitly stated meaning in the navigational cells (lower row)

the cells containing the variable. Data about each cell in our case contains cell content, a content of its header, stub and super-row number, cell's role and the position of the cell in the table gird. The content of the cell and its navigational areas are stemmed using Porter stemmer, tokenized, and the bag-of-words methodology was used. We modelled problem as a classification task, in which if the cell contains variable the classification returns positive class and negative class otherwise.

### 3.4.7 Pattern analysis and value extraction

Once in this step, the method knows that cell contains the variable. However, information can be presented in various formats. In this step, the content of the cell is analysed and the value is searched based on the number of possible information presentation patterns. For the case of numeric information, these patterns can be crafted by using regular expressions. These patterns also may have a logic on how to translate the presented information to the extraction template. For example, if the presented value for patient age is "18.3 (16–27)", the logic bound to the pattern should be able to deduce that the number 18.3 is a mean value, while 16–27 is a range, where the first value is minimum and the second is maximum patient age. After the pattern analysis is performed, the value populates the information extraction template.

Before the information is extracted, the syntax of the selected cell is analysed against a set of syntactic patterns. These patterns are pre-defined to inform the method how to disentangle and interpret the content of the cell. Cells often contain complex value presentation patterns and represent multiple information (see example in Fig. 4). Authors usually use same or similar value presentation patterns to present similar information (e.g. *variable value, mean, standard deviation, percentage, alternative values,* etc.). Patterns provide the way to extract atomic information and to provide the value presentation semantics. For example, if the value is presented as $16 \pm 3.2$, it is possible to determine that the first value is mean or median, while the second is the standard deviation or standard error. In order to exactly specify the semantics of each value component, the methodology looks at the related access cells. Based on these patterns, information is extracted and stored in the database.

## 3.5 Defining rules for information extraction

### 3.5.1 Cell selection using lexical and semantic rules

In a heuristic-based approach, selecting cells in which a variable and its value are presented is done using a defined set of lexical and semantic cues.

Lexical cues are defined as a set of words in white list and blacklist. A target, functional location of the cue (header, stub, super-row, target) is defined for both lists. In other words, a definition is provided as to whether the cue should be searched for in the header, stub or super-row or in the target cell.

Table cells are iterated and tested against the defined lexical rules. The presence of the cue from the white list signals that the target cell potentially contains a value for the variable of interest. The cell is then tested against cues from the blacklist. If the cell or its navigational cells contain cues from the blacklist, the selection is discarded.

Semantic cues are defined similarly to lexical cues. However, instead of words or phrases that are searched for, we use annotations. Annotations can be searched for in headers, stubs, super-rows of the target cell or in the target cell itself. Again, an annotations white list and blacklist is used. The method uses two layers of annotations: annotation id and annotation description. In the case of UMLS annotations, annotation ids were UMLS concept ids, while annotation descriptions were semantic types. Therefore, it was possible to create white lists and blacklists consisting of UMLS concept ids and semantic types for the UMLS annotated data. This method iterates through table cells, selects cells using signals from white lists and discards cells containing cues from the blacklist. It is also possible to combine lexical and semantic cues while creating cue lists (black or white). The user has to input a list of words/concepts, one per line, where lexical cues are preceded with the sequence "[word]:", while concepts are preceded with the sequence "[annID]:" and UMLS semantic types are preceded with the sequence "[annType]:". For an example see Fig. 5.

In this step, the method also selects the unit and context for the numerical variables. A set of possible units for the given variable has to be defined as well as the default value. The method searches the cell and its navigational areas (header, stub, super-row) for a mention of the unit. If a unit is found, it is extracted and if not, the default unit is used.

In our method, the context is extracted as the concatenated value of navigational cells relevant to the target cell that did not contain cues from the white list.

### 3.5.2 Syntactic rules and syntactic processing

The role of syntactic processing is to analyse the content of the selected cell with the value, disentangle the value and identify its components (populating *Value Component* from the extraction template). For example, the syntactic processing reveals whether the extracted value is the mean, median, standard deviation, range, percentage, etc.

The value patterns are common for certain types of information. For example, age, BMI, FEVl and many other variables present overall statistics for certain population (*average, mean, standard deviation, range*). If the rules are developed for one variable, they can be reused for others. In this way, it is possible to create a library of common value presentation patterns. Examples of common numerical presentation patterns are presented in Table 3.

Syntactic processing is performed using a rule-based methodology. We propose a method for describing syntactic value pattern disentangling rules. The description methodology uses regular expressions for disentangling cell content. Syntactic rules map values to their descriptions.

A definition of a syntactic rule contains three components: (1) the rule's name, (2) the rule's regular expression and (3) a set of semantic assignments (descriptions) for each component of the regular expression.

Value components (*e.g. mean, standard deviation, range-min, range-max, etc.*) can be assigned to each regular expression component. The aim of syntactic processing is to assign semantics to each value component based on the value presentation pattern and cues in the pattern and navigational cells related to the cell. Therefore, a set of possible but distinct semantic assignments can be listed with the regular expression defining the rule and giving possible meanings to each extracted value. Often, a value's semantics can be induced from the value presentation pattern. For example, if a table's cell contains BMI values of *20–37*, it is likely that the value is the range, with a minimum value of 20 and a maximum value of 37. However, for some value presentation patterns, additional information in the navigational part of the table is necessary. One example is a pattern like "$16 \pm 4$". The first value could be either the mean or median. The navigational cells' content for these data cells will determine through mention, whether the value is mean or median. If the definition is not mentioned, a default assignment of the value's meaning can be used by applying the most common one. In other cases, multiple values are presented with the explicitly described semantics of each value part in navigational areas. For example, if the gender value is presented as *15:14*, navigational cell's would describe which value presents the number of male participants and which one is the number of female participants. We allow for each extracted regular expression group to define a set of keywords or synonyms with their order of appearance that is looked for in navigational areas. The semantic assignment contains a group number, ordered groups of keywords (or synonyms) and the semantic assignment. Each keyword group is a comma-separated list of strings. A semantic assignment value is separated by the arrow symbol ($\rightarrow$). Figure 6 provides an example of a rule that can disentangle a pattern such as *"15:14"* for gender. According to the rule, in case any cue from the list linked to the number of male participant variable (*male, m, Male, M, men, males, Males*) appears before any cue linked to the number of female participants variable (*female, f, fem, Fem, women, Women, females, Females*), the first value is associated with the number of male participant variable. In case a cue from the list linked to female participants is appearing first, the first value is the number of female participants, while the second value is the number of male participants. In case none of the cues appear, as default, rule assigns the first value to the number of male participants, while the second is assigned to the number of female participants.

**Fig. 5** Example of one syntactic rule with its semantics for extracting gender distribution of the participants

**Table 3** Examples of common syntactic patterns and variables that are often represented by them

| Pattern | Presentation examples | Variables |
|---|---|---|
| Single value | 65 | Number of patients, number of people with certain adverse event, etc. |
| Floating point value | 0.05 | $P$ value |
| Aggregate statistical value | $18 \pm 2$ | Age, FEV1, PEF, BMI |
| | 12–18 | Weight, height |
| | 12.1 (2.4) | Number of patients in cohort |
| | $18 \pm 2$ (15–20) | |
| Alternatives | 12/17 | Gender distribution, blood pressure |
| Percentage | 18 (55%) | Gender distribution |
| | 55% | Percentage of people with certain effect |

Another example of the rule definition for statistical values (range, mean, median and standard deviation) can be seen in Fig. 7.

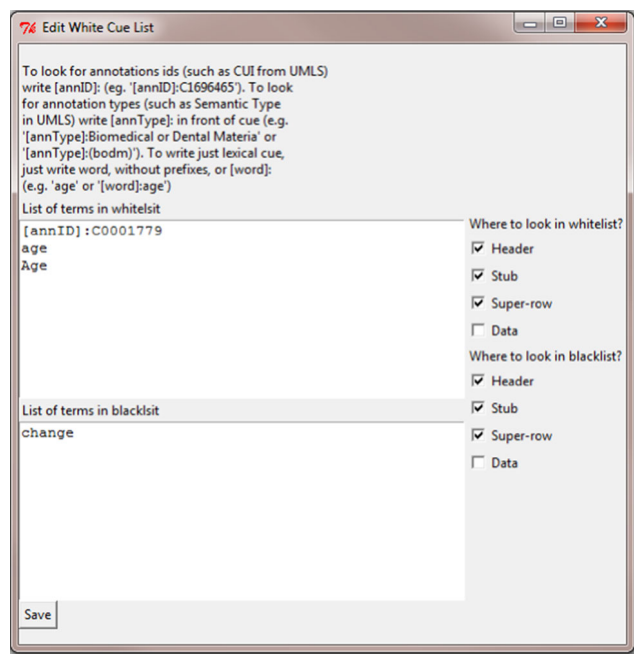

**Fig. 6** Example of one lexical and semantic rule definition in TabInOut wizard

```
+GetMean1
(\d+\.*\d*)[?    ]*[\-------,]+(\d+\.*\d*)[ (]*
(\d+\.*\d*)[?    ]*[±][?    ]*(\d+\.*\d*)[)]*
1->range_min
2->range_max
3:median,Median->median
3->mean
4->SD
```

**Fig. 7** Example of one syntactic rule with its semantics for extracting statistical values

In the case of the cell with the content "$12-18(16 \pm 4)$", the rule from Fig. 7 would say that 12 is minimum value of a range, 18 is a maximum value, 16 is mean or median (in case median is mentioned in stub or header of the cell) and number 4 is standard deviation.

```
▼<table-wrap id="tbl1" position="float">
    <label>Table 1</label>
  ▼<caption>
      <p content-type="table-title">Baseline demographic and disease characteristics</p>
    </caption>
  ▼<table border="1" frame="hsides" rules="groups" width="85%">
    ▼<colgroup>
        <col align="left"/>
        <col align="char" char="."/>
      </colgroup>
    ▼<thead valign="bottom">
      ▼<tr>
          <th align="left" charoff="50" valign="top">Number of patients enrolled</th>
          <th align="center" char="." charoff="50" valign="top">21</th>
        </tr>
      </thead>
    ▼<tbody valign="top">
      ▼<tr>
          <td align="left" charoff="50" valign="top">Median age (range)</td>
          <td align="center" char="." charoff="50" valign="top">57 (36-2) years</td>
        </tr>
      ▼<tr>
        ▼<td align="left" charoff="50" valign="top">
            <italic>Sex</italic>
          </td>
          <td align="char" char="." charoff="50" valign="top"></td>
        </tr>
      ▼<tr>
          <td align="left" charoff="50" valign="top">Male</td>
          <td align="center" char="." charoff="50" valign="top">15</td>
        </tr>
      ▼<tr>
          <td align="left" charoff="50" valign="top">Female</td>
          <td align="center" char="." charoff="50" valign="top">6</td>
        </tr>
      ▼<tr>  • • •
```

**Table 1**

Baseline demographic and disease characteristics

| Number of patients enrolled | 21 |
|---|---|
| Median age (range) | 57 (36–2) years |
| Sex | |
|     Male | 15 |
|     Female | 6 |
| Performance status | |
|     0 | 5 |
|     1 | 13 |
|     2 | 3 |

a

Brain metastasis previously resected.

**Fig. 8** Example of table in PMC XML format and its visualisation

For categorical variables, syntactic analysis depends on the user's definition of possible categories for that variable. Patterns can be defined as possible representations of the given category (e.g. synonyms) that algorithm matches and extracts from the cells' content.

For textual variables, syntactic analysis has to be complemented with further lexical and semantic analysis in order to extract more granular information from the cell content. However, this is outside the scope of this paper.

## 4 Applications and results

### 4.1 Dataset

The dataset was created by mapping public PMC[4] articles from 2014 with MEDLINE[5] citations that contained word "clinical" in publication type. This dataset contains 6,109 articles with 14,009 tables. The example of a table in PMC XML format and its visualisation can be seen in Fig. 8. The dataset that was used for evaluation of PMC data can be found at https://data.mendeley.com/datasets/wk53twxddf/1.

### 4.2 Functional and structural table analysis

On evaluation set, our method performs functional analysis with a precision of 0.9425, recall of 0.9428 and $F1$-score of 0.9426. Relationships between cells were recognized with a precision of 0.9238, recall of 0.9744 and $F1$-score of 0.9484 [30].

### 4.3 Pragmatic table analysis

The training set for pragmatic classification contained 186 tables labelled as baseline characteristic tables, 60 inclusion/exclusion criteria tables, 239 adverse event tables and 153 classified as others. We extracted features as we described in Sect. 3.4.5. A number of machine learning algorithms were tested, including Naive Bayes, SVM, decision trees, random tree and random forest. The evaluation was performed using the 10-fold cross-validation. The results of the pragmatic classification experiments are presented in Table 4.

Caption and stubs are good features for the pragmatic analysis. This is expected as caption's purpose is to describe the table and its content. Caption often describes what information is grouped together. On the other hand, stubs contain concept names that, when grouped together, can help identify a table's pragmatic type. Other features were not as successful in predicting pragmatic type. The header information usually contains names of clinical arms or drugs, which is not

---

**Table 4** Weighted averages for all classes of the pragmatic classification using different content feature sets (each content feature separately and all features combined)

| Algorithm | Precision | Recall | *F*-Score |
|---|---|---|---|
| Numeric features | | | |
| Naive Bayes | 0.569 | 0.601 | 0.553 |
| Bayesian Networks | 0.491 | 0.552 | 0.499 |
| SVM | 0.475 | 0.559 | 0.493 |
| C4.5 decision trees | 0.498 | 0.503 | 0.500 |
| Random forests | 0.558 | 0.580 | 0.562 |
| Caption text | | | |
| Naive Bayes | 0.901 | 0.902 | 0.901 |
| Bayesian networks | 0.907 | 0.905 | 0.906 |
| SVM | 0.930 | 0.930 | 0.930 |
| C4.5 Decision tree | 0.926 | 0.925 | 0.926 |
| Random forests | 0.889 | 0.889 | 0.888 |
| Header text | | | |
| Naive Bayes | 0.687 | 0.654 | 0.660 |
| Bayesian Networks | 0.682 | 0.634 | 0.642 |
| SVM | 0.648 | 0.631 | 0.635 |
| C4.5 Decision tree | 0.659 | 0.612 | 0.620 |
| Random forests | 0.646 | 0.628 | 0.618 |
| Stub text | | | |
| Naive Bayes | 0.821 | 0.796 | 0.801 |
| Bayesian networks | 0.841 | 0.802 | 0.807 |
| SVM | 0.808 | 0.772 | 0.776 |
| C4.5 Decision tree | 0.821 | 0.779 | 0.783 |
| Random forests | 0.803 | 0.776 | 0.780 |
| Super-row text | | | |
| Naive Bayes | 0.568 | 0.477 | 0.461 |
| Bayesian networks | 0.696 | 0.440 | 0.490 |
| SVM | 0.526 | 0.448 | 0.373 |
| C4.5 Decision tree | 0.691 | 0.508 | 0.476 |
| Random forests | 0.694 | 0.537 | 0.514 |
| Data cell content | | | |
| Naive Bayes | 0.573 | 0.556 | 0.551 |
| Bayesian networks | 0.572 | 0.568 | 0.567 |
| SVM | 0.604 | 0.586 | 0.587 |
| C4.5 Decision tree | 0.560 | 0.551 | 0.551 |
| Random forests | 0.603 | 0.592 | 0.587 |
| Referring sentence | | | |
| Naive Bayes | 0.726 | 0.590 | 0.618 |
| Bayesian networks | 0.698 | 0.618 | 0.625 |
| SVM | 0.682 | 0.625 | 0.626 |
| C4.5 Decision tree | 0.630 | 0.575 | 0.573 |
| Random forests | 0.675 | 0.622 | 0.617 |
| Combined content features | | | |
| Naive Bayes | 0.873 | 0.871 | 0.872 |
| Bayesian networks | 0.865 | 0.864 | 0.864 |

**Table 4** continued

| Algorithm | Precision | Recall | *F*-Score |
|---|---|---|---|
| SVM | 0.915 | 0.914 | 0.914 |
| C4.5 Decision tree | 0.883 | 0.880 | 0.881 |
| Random forests | 0.917 | 0.915 | 0.916 |

The evaluation was done using 10-fold cross-validation

**Table 5** Results of the four-class pragmatic classification experiments on the PMC clinical trial tables using combination of quantitative (number of cells, number of rows, number of columns, percentage of empty cells, percentage of numeric cells, percentage of text cells) and content features (content of the cell, header, stub, caption, footer)

| Algorithm | Precision | Recall | *F*-Score |
|---|---|---|---|
| Naive Bayes | 0.943 | 0.943 | 0.943 |
| Bayesian Networks | 0.938 | 0.939 | 0.938 |
| C4.5 decision trees | 0.944 | 0.945 | 0.944 |
| Random tree | 0.905 | 0.903 | 0.904 |
| Random forests | 0.948 | 0.948 | 0.948 |
| SVM | 0.967 | 0.966 | 0.966 |

Training and evaluation was performed using the 10-fold cross-validation on 186 "baseline characteristic", 60 "inclusion/exclusion", 239 "adverse event" and 153 "other" tables as relevant for the pragmatic analysis, as it is defined in this case. When a header is used only as a content feature, it achieves *F*1-scores between 0.618 and 0.66 depending on the algorithm used. The data cells' content presents concept values, but little can be concluded from these values without the descriptions from the table's navigational areas. Expectedly, *F*1-scores for data cells, as the only content feature, are in a range of 0.551–0.587. Referring sentences sometimes describe tables but more often they analyse or compare the results or just refer to the table (e.g. "See table X"). From the analysis, comparison or reference often cannot infer the purpose of the table without additional information. Referring sentences' *F*1-scores range between 0.573 and 0.626. Super-rows can be as good classification feature as stubs (of which they are a part); however, many tables do not contain super-rows. Therefore, when only super-rows are used as content features, the *F*1-score produces a range of 0.373–0.490.

The final classifier used some of the content features, such as the stub, caption, header content and quantitative features, such as the number of columns, number of rows and order of the table in the article. We used an SVM classifier with mentioned features that had a precision of 0.967 and recall of 0.966 when evaluated on the data set using 10-fold cross-validation. The results of the final classifier combining quantitative and content features are presented in Table 5.

The dataset of clinical trial papers from PMC has 6558 articles containing 12787 tables. According to the pragmatic

**Table 6**  Pragmatic distribution of tables

| Table type | Number |
| --- | --- |
| Baseline characteristics | 2803 (21.92%) |
| Adverse events | 633 (4.95%) |
| Inclusion/exclusion | 82 (0.47%) |
| Other | 9291 (72.66%) |

**Table 8**  Results of information extraction for the number of patients

| | Precision | Recall | $F$-Score |
| --- | --- | --- | --- |
| Training | 0.900 | 0.839 | 0.868 |
| Testing | 0.894 | 0.791 | 0.839 |

table classifier, the distribution of tables, according to our pragmatic classification model, is presented in Table 6.

We have also tested an approach in which we defined broader pragmatic classes (experimental settings, experimental results and supporting knowledge, that would include literature review, definitions of scales, terms or examples). With this approach, the best performance of the machine learning algorithm was around 0.85 $F$1-score or about 10% worse than with more specific pragmatic classes (baseline characteristic, adverse events, inclusion/exclusion).

### 4.4 Rule-based information extraction

We implemented and evaluated a rule-based methodology for extracting the total number of patient in a clinical trial, statistics about the age of the patients (mean, standard deviation and range) and names of adverse events.

**Number of patients extraction.**  We firstly checked caption of each table for pattern stating with a number followed by a trigger word in its vicinity (patients, subjects, individuals, participants, etc.). If the pattern is found, the number is extracted as a total number of trial participants. We also select cells containing trigger words and phrases in their stub. The header usually represent arm or treatment group, which maps to *context* in our extraction template (Sect. 3.1). The number in the cell is extracted as a candidate for the number of participants in that group. Candidates are checked against the blacklist of cues that would determine that the value is not the number of participants (e.g. $P$ value, %, mean, median). If the content of header, stub, and the cell do not contain these words, the value is extracted. We also select header cells containing the letter "$n$" and the number ($n = 19$). The number next to letter "$n$" is extracted and the content of the cell without the expression is considered a participant group name (source). Example table presenting the number of patients in the header and caption is presented in Table 7. We created the rules based on randomly selected 100 tables, with baseline characteristic pragmatic class, extracted from clinical trial publications in PMC (training set). Evaluation (testing) set contained another 100 randomly selected baseline characteristic tables from clinical trial papers. The results of the manual evaluation of information extraction of the number of the patients can be seen in Table 8.

The algorithm extracted 4355 values as the number of patients with mentioned precision and recall from 6558 documents. Since some tables presented the number of patients per clinical trial arm or participant group, there were only 1699 documents (26%) presenting the number of patients in tables. Our initial hypothesis was that the majority of clinical trial documents should report the number of patients. In order to examine why our method extracted the number of patients only from 26% of documents, we examined a sample of 25 documents (containing 98 tables) from which the number of patients was not extracted and found following reasons:

– Document contained no tables
– There was no baseline characteristic table, and the number of patients was not presented in any table (may have been present in the text)
– There was no baseline characteristic table; however, the number of patients was presented in some other table (e.g. results, referral question)
– Table reported results per person. In this case, the number of patients can be calculated as the number of data rows

**Table 7**  Example table presenting the number of patients in caption and header (PMC270000)

| Parameter | Bravelle® ($n = 120$) | Follistim® ($n = 118$) | $P$ value |
| --- | --- | --- | --- |
| Age (years) | 32.0 ± 3.9 | 32.5 ± 3.7 | 0.330 |
| Weight (lbs.) | 137.1 ± 21.4 | 145.8 ± 27.8 | 0.008 |
| Body mass index (kg/m$^2$) | 23.3 ± 3.5 | 24.5 ± 4.0 | 0.021 |
| Serum FSH (mIU/mL) | 6.3 ± 2.0 | 6.8 ± 2.1 | 0.077 |
| Serum LH (mIU/mL) | 5.0 ± 2.4 | 4.6 ± 1.9 | 0.145 |
| Serum E2 (pg/mL) | 43.1 ± 21.4 | 40.9 ± 20.9 | 0.420 |

Baseline demographic characteristics (prior to leuprolide acetate) of the **120 patients** who received Bravelle® and the **118 patients** who received Follistim®

**Table 9** Results of information extraction for age of patients, including mean, standard deviation and range

|  | Precision | Recall | $F$-Score |
| --- | --- | --- | --- |
| Training | 0.806 | 0.895 | 0.848 |
| Testing | 0.788 | 0.872 | 0.828 |

in a given table. However, our method was looking for the cumulative number of patients.

- Baseline characteristic table does not report the number of patients. It is mentioned in the text or it is possible to calculate it from gender distribution.
- Error of either pragmatic classifier or rule did not contain the right cue (e.g. infants, smokers).

The errors in extraction appeared because we did not compile exhaustive lexical cue list. This caused both false positives and false negatives. We missed some rare cues that represented the number of participants. Also, certain words that should be part of the blacklist were missed (i.e. "number of patients excluded"—the word "excluded" in a given example). Also, the use of abbreviations, especially nonstandard ones, caused false negatives (Num. patients, No. patients, N patients, # patients, with possible changes in word order). Most of the cues or abbreviations that we did not capture were not present in the training set, while some were specific to a given paper.

**Patients' age extraction** A similar methodology was applied for extraction of the statistics about patients' age. Algorithm selected candidate cells based on lexical cues appearance in stub and super-row. These candidates were filtered through a lexical cue blacklist specifically designed for this. Once the right candidates were selected, we extracted numbers against a number of regular expression patterns (mean $\pm$ standard deviation, min-max, mean (min-max), etc.). Age may be presented in several units (years, months, weeks, days). We checked stub and header value for the appearance of this cues. In case some of these units are presented in the navigational area, that unit is recorded in the unit field of our template. Otherwise, "year" is recorded as default unit. We evaluated extraction of patient age (mean, standard deviation and range) using the same training and testing dataset as for extraction of the patient number. The results can be seen in Table 9.

Age was presented in 1944 documents (30%). Our method extracted 13182 values as patient age. Our algorithm extracted 6125 instances of mean age, 2475 instances of standard deviation and 2291 instances of age ranges. Compared to the number of patients, age is more commonly presented in tables and with more complexity (mean, standard deviations, range).

We encountered a couple of tables that were not recognized as baseline characteristic tables. One of the tables contained cue "age" in unexpected context (HT age—hormone therapy age). In two tables, we had value patterns that were not expected (they were not in training set), so our algorithm was able to extract only mean value and missed standard deviations that were presented in the table. Four tables presented age groups with a number of trial participants in each of the group. The algorithm misinterpreted these numbers as mean ages of participants. In three tables, super-row or second header of multi-table were not recognized correctly which led to false negatives.

Matching patterns and extracting the right values once the value is recognized is an important part of the rule-based approach. We have evaluated the performance of pattern extraction on the extraction of patient age. The statistics about patient age can be presented as mean value, its standard deviation and the range of ages. However, even these three values can be combined and presented in various syntactically different formats. During the evaluation, we were examining only the cells that were recognized as a cell containing the age of patients correctly and we were evaluating the performance of pattern matching.

The patterns were matched and extracted with a precision of 99.4%, recall of 95.75% and an F1-score of 97.54%. The evaluation is done on statistical patterns representing the age of the patients (means, medians, ranges and standard deviations were extracted). In the testing set, there were several patterns that did not appear in training data that included some special characters (central dot ($\cdot$) instead of dot(.)) and one new presentation pattern. However, pattern matching is reliable, accurate and reusable. Once developed for a certain type of presentations or value group (such as cumulative statistical data), it can be reused for other information classes that present information in the same manner.

**Adverse event names extraction** For extracting adverse reaction names, we performed slightly different approach. Firstly, we selected tables that pragmatic classification classified as the ones containing adverse events. We used the UMLS semantic type annotations of the content of the cells in order to recognize whether the certain column contains adverse events. Firstly, we annotated the content of cells with semantic types using MetaMap. For each column, our method checks whether cells contain phrases annotated as "Sign or Symptom" or "Disease or Syndrome". In case the majority of cells in the same columns contain this annotation, the content of all cells in that column, except header, is extracted as adverse event names. We performed an evaluation of detecting names of adverse events over 35 documents in training and 35 documents in the testing set. Results are presented in Table 10.

**Table 10** Results of information extraction for adverse events

|  | Precision | Recall | $F$-Score |
|---|---|---|---|
| Training | 0.945 | 0.906 | 0.925 |
| Testing | 0.883 | 0.962 | 0.921 |

MetaMap annotated 7701 instances of adverse events, while 4974 adverse event instances that were not annotated were extracted with presented accuracy from 6558 clinical articles with 12,787 tables.

The extraction of adverse event names gave better performance than the extraction of numeric values (number of patient and age). This is mainly due to UMLS annotations that could be utilized for adverse events. Errors appeared in columns that contained mixed content, among which were adverse events. Also, one table listed diseases, which were by our approach recognized as adverse events.

## 4.5 Machine learning-based information extraction

Another approach to extract information from tables is by using machine learning in order to detect cells containing variables of interest. The information is then extracted using patterns, similarly to the rule-based approach. We have implemented detection of cells containing a number of patients, information about the age of patients and gender distribution. The aim was to make a machine learn cues and usual patterns for presenting the variable or its value. In order to make it easier for the machine learning algorithm to learn the presentation patterns of numeric values, we changed numeric symbols to "$\times$" symbol.

We created a training dataset using 100 randomly selected baseline characteristic tables from PMC. For a number of patients, there were 147 positively labelled cells. The number of cells presenting age of the patients was 272, while there were 204 cells presenting gender. The cells that were in positive class for each of the variable (number of patients, age, gender) were cells presented numerical values for these classes in a given table. The dataset was highly imbalanced since the whole dataset contained 13,610 cells. We applied three machine learning techniques. In the first one, we balanced the dataset for each learning task, so it contained the same number of negatively labelled cells as positively labelled ones (under-sampling). For this technique, we performed learning on under-sampled data, while we evaluated it on the large unseen dataset (containing 7261 cells). The second approach consisted of learning from the unbalanced dataset. In the third approach, we used cost-sensitive classification and experimentally adjusted the weights for the best performance. For the second and third approaches, we performed 10-fold cross-validation in Weka. The results are presented in Tables 11, 12 and 13.

**Table 11** Results of the patient number information extraction based on machine learning

| Algorithm | Under-sampled (147 instances of each class) | | | | Whole unbalanced dataset | | | | Cost-sensitive classification | | | |
|---|---|---|---|---|---|---|---|---|---|---|---|---|
| | Precision | Recall | $F$-Score | Accuracy | Precision | Recall | $F$-Score | Accuracy | Precision | Recall | $F$-Score | Accuracy |
| Naive Bayes | 0.054 | 0.952 | 0.103 | 0.821 | 0.173 | 0.701 | 0.277 | 0.960 | 0.266 | 0.483 | 0.343 | 0.980 |
| Bayesian Nets | 0.101 | 0.912 | 0.182 | 0.911 | 0.292 | 0.517 | 0.373 | 0.981 | 0.512 | 0.422 | 0.463 | 0.989 |
| C4.5 dec. trees | 0.070 | 0.905 | 0.130 | 0.869 | 0.893 | 0.510 | 0.649 | 0.994 | 0.714 | 0.782 | 0.747 | 0.994 |
| Random tree | 0.066 | 0.585 | 0.119 | 0.906 | 0.580 | 0.544 | 0.561 | 0.991 | 0.573 | 0.585 | 0.579 | 0.991 |
| Random forests | 0.214 | 0.932 | 0.348 | 0.962 | 0.935 | 0.490 | 0.643 | 0.994 | 0.797 | 0.667 | 0.726 | 0.995 |
| SVM | 0.085 | 0.918 | 0.155 | 0.892 | 0.850 | 0.463 | 0.599 | 0.993 | 0.754 | 0.626 | 0.684 | 0.994 |

**Table 12** Results of the age of patients (cumulative statistical values such as mean, standard deviation and range) information extraction based on machine learning

| Algorithm | Under-sampled (272 instances of each class) | | | | Whole unbalanced dataset | | | | Cost-sensitive classification | | | |
|---|---|---|---|---|---|---|---|---|---|---|---|---|
| | Precision | Recall | F-Score | Accuracy | Precision | Recall | F-Score | Accuracy | Precision | Recall | F-Score | Accuracy |
| Naïve Bayes | 0.089 | 0.930 | 0.162 | 0.879 | 0.205 | 0.819 | 0.327 | 0.957 | 0.254 | 0.754 | 0.381 | 0.969 |
| Bayesian Nets | 0.128 | 0.918 | 0.224 | 0.920 | 0.419 | 0.743 | 0.536 | 0.984 | 0.504 | 0.684 | 0.581 | 0.987 |
| C4.5 dec. trees | 0.092 | 0.795 | 0.165 | 0.899 | 0.886 | 0.591 | 0.709 | 0.994 | 0.783 | 0.801 | 0.792 | 0.995 |
| Random tree | 0.074 | 0.871 | 0.136 | 0.900 | 0.628 | 0.573 | 0.573 | 0.990 | 0.628 | 0.573 | 0.599 | 0.990 |
| Random forests | 0.213 | 0.947 | 0.348 | 0.963 | 0.945 | 0.503 | 0.656 | 0.993 | 0.883 | 0.661 | 0.756 | 0.995 |
| SVM with SMO | 0.180 | 0.614 | 0.278 | 0.963 | 0.955 | 0.743 | 0.836 | 0.996 | 0.895 | 0.801 | 0.846 | 0.996 |

**Table 13** Results of the cell detection using machine learning for target class of gender distribution

| Algorithm | Under-sampled (204 instances of each class) | | | | Whole unbalanced dataset | | | | Cost-sensitive classification | | | |
|---|---|---|---|---|---|---|---|---|---|---|---|---|
| | Precision | Recall | F-Score | Accuracy | Precision | Recall | F-Score | Accuracy | Precision | Recall | F-Score | Accuracy |
| Naïve Bayes | 0.075 | 0.929 | 0.139 | 0.834 | 0.155 | 0.675 | 0.252 | 0.942 | 0.167 | 0.584 | 0.260 | 0.952 |
| Bayesian Nets | 0.099 | 0.934 | 0.179 | 0.876 | 0.475 | 0.584 | 0.524 | 0.985 | 0.813 | 0.528 | 0.640 | 0.991 |
| C4.5 dec. trees | 0.119 | 0.929 | 0.210 | 0.899 | 0.912 | 0.685 | 0.783 | 0.994 | 0.839 | 0.766 | 0.801 | 0.994 |
| Random tree | 0.081 | 0.959 | 0.150 | 0.843 | 0.739 | 0.746 | 0.742 | 0.992 | 0.739 | 0.746 | 0.742 | 0.992 |
| Random forests | 0.155 | 0.990 | 0.218 | 0.922 | 0.953 | 0.624 | 0.755 | 0.994 | 0.893 | 0.807 | 0.848 | 0.996 |
| SVM with SMO | 0.122 | 0.909 | 0.909 | 0.897 | 0.903 | 0.756 | 0.823 | 0.995 | 0.833 | 0.812 | 0.823 | 0.995 |

Precision, recall and $F1$-score presented in the tables are measured on positive class (number of patients, age, gender). Since the data is unbalanced, the weighted average does not present representable information.

The dataset contains a small number of instances of the positive class. Because of the small number of positive instances, balancing data by under-sampling is not performing well. Most machine learning algorithms rely on probabilistic distribution of classes and assume the same costs for misclassification of classes [16]. However, if the data represent a realistic distribution of classes, some of the algorithms are able to cope with data relatively well. As it can be seen from the tables, results from learning from the whole dataset for some algorithms, such as decision trees, random forests and SVM, are much better than with under-sampled data. By using cost-sensitive classification and assign larger costs to positive class than to negative, it is possible to improve these results. We managed to improve the $F1$-scores by almost 10% (see Tables 11, 12, 13).

However, when the machine learning approach is compared with the rule-based approach, it can be seen that a simple rule-based approach with a white list and blacklist of lexical cues was performing with similar or even better $F1$-scores. Development of machine learning model is more complex and time-consuming than crafting white and blacklists because usually; it is necessary to annotate several thousand cells and perform a number of experiments to find the most suitable costs. Within tables, it is easier than in free text to craft lexical cues and rules for information extraction, after the table is disentangled and the scope is narrowed by pragmatic classification. Also, rules can be efficiently improved at any time by adding or modifying a set of rules. Improvement in machine learning approach would require additional annotations and generate a completely new model.

# 5 Generalizability case study

In order to evaluate the generalizability of the framework, we designed a case study on extracting drug–drug interactions from tables in Structured Product Labels that are available through DailyMed,[6] a National Library of Medicine's database of approved drug labels in the USA.

Structured Product Labels (SPL) is a document markup standard (a variant of XML) approved by Health Level Seven (HL7) and adopted by the United States' FDA as a mechanism for exchanging product and facility information.[7] SPL documents annotate certain information, such as drug name, ingredient substances or manufacturer. However, they also contain a number of sections with text, figures and tables. Section names and topics are prescribed by the FDA and annotated with Logical Observation Identifiers Names and Codes (LOINC).

This case study was designed to prove that it is possible with minimal modification to transfer methodology to the different dataset with different markup (DailyMed compared to PMC), different domain (drug labels compared to clinical trial) and slightly different task (relation extraction compared to single information extraction). We will evaluate each step of the methodology and discuss any modification that is necessary.

## 5.1 Document reading and table detection

As we focus on XML format, and documents from the DailyMed database are in XML, it is trivial to detect tables. However, XML tags and emphasis features are not the same. Therefore, in TableDisentangler, a new document reader had to be developed for this type of documents. For each document format, our class model requires cells to be imputed into a grid structure, together with their features (whether there is bold text, italic text, whether cell spans horizontally, vertically, for how many cells it spans). This grid is then used by the other parts of the methodology (functional processing, structural processing, pragmatic analysis, table annotation, cell selection and syntactic analysis).

We downloaded structured product labels for all 30,409 prescription drug products as of January 1, 2016, from DailyMed. The full set of SPLs was reduced to a subset of SPLs identified as having at least one table in the Drug Interaction section (section coded with LOINC 34073-7). The data contained 16,211 tables from 1161 SPL documents. However, only 1530 tables contained information about drug–drug interactions. (They were presented in the drug interaction section.) These SPLs were used as input in TableDisentangler, which parsed and analysed the table content and assigned functional roles and structural relationships to individual cells and annotated the contents of each cell.

## 5.2 Functional and structural processing

The functional analysis determines each cell's functions within each table. Cells are identified as table header, row header, super-row or a data cell.

The TableDisentangler methodology is primarily based on emphasis features. However, the DailyMed dataset does not follow the same emphasis rules, especially for headers. Headers are not divided by horizontal lines and are often not marked with *thead* tags. Approximately 46% of tables presenting drug–drug interactions (565 tables) did not have labelled headers. The caption can be also presented inside

| Table 7: Summary of AED Interactions with Oxcarbazepine | | | | |
|---|---|---|---|---|
| AED Coadministered | Dose of AED (mg/day) | Oxcarbazepine dose (mg/day) | Influence of Oxcarbazepine on AED Concentration (Mean change, 90% Confidence Interval) | Influence of AED on MHD Concentration (Mean change, 90% Confidence Interval) |
| Carbamazepine | 400 to 2000 | 900 | nc[1] | 40% decrease [CI: 17% decrease, 57% decrease] |
| Phenobarbital | 100 to 150 | 600 to 1800 | 14% increase [CI: 2% increase, 24% increase] | 25% decrease [CI: 12% decrease, 51% decrease] |
| Phenytoin | 250 to 500 | 600 to 1800 >1200 to 2400 | nc[1,2] up to 40% increase[3] [CI: 12% increase, 60% increase] | 30% decrease [CI: 3% decrease, 48% decrease] |
| Valproic acid | 400 to 2800 | 600 to 1800 | nc[1] | 18% decrease [CI: 13% decrease, 40% decrease] |

[1]nc denotes a mean change of less than 10%
[2]Pediatrics
[3]Mean increase in adults at high oxcarbazepine doses

**Fig. 9** Example of a table in which both caption and footer are inside the table cells (DailyMed setID: 524c025b-809b-440f-a756-e3518d7c92db)

**Fig. 10** Workflow of the modified methodology for functional and structural analysis of DailyMed documents

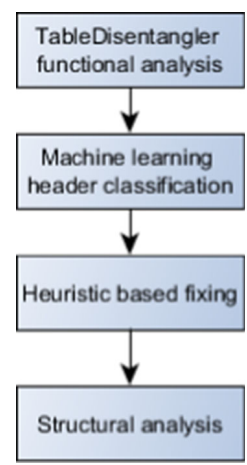

*thead* tags, while the actual table header is below, in the body of the table. We queried cell content for cues that indicate caption (word "Table" followed by the number) and found 136 tables containing caption in one of the cells, often labelled as header (see example in Fig. 9).

We evaluated the performance of the original TableDisentangler methodology for functional analysis. We randomly selected 20 tables for evaluation and inspected them manually. Cells were considered true positives if their function was correctly annotated. If the correct function was not annotated, it was counted as false negative, while if the cell was annotated with incorrect functional annotation, it was considered false positive. The results are presented in Table 14. For most of the functional classes, the methodology performed well, apart from detecting headers. The precision of header detection was 0.61, while the recall was 0.65. Headers are important for extracting drug–drug interactions since header labels can be used efficiently to craft extraction rules. The evaluation showed that headers have to be treated differently for the DailyMed dataset by taking lexical and semantic cues into account.

In order to improve the performance of header detection, we developed a hybrid methodology consisting of a machine learning model and heuristics (see workflow diagram in Fig. 10). As only the header detection in DailyMed documents is performing with low scores, we used the methodology described in [30] for classifying stubs and super-rows. Firstly, TableDisentangler with the standard functional analysis methodology is executed. Secondly, we applied a machine learning algorithm to classify header cells based on their content. In order to train the algorithm, we randomly selected 1,000 headers labelled by TableDisentangler from drug–drug interaction tables. They were manually reviewed and relabelled by a final year pharmacology student. For training, we selected 823 labels, 329 headers and 494 non-headers. The performance of the machine learning method using 10-fold cross-validation on the described dataset is presented in Table 15. Thirdly, we used heuristics to post-process functional annotations. We assume that all cells of a certain row have to be either in the header row or outside it. Therefore, if the majority of the cells in some row are classified as headers, then all the other cells in that rows are also annotated as part of the header. If a minority of the cells in the row is classified as headers, their annotations are fixed to data cells. Based on experimental experience, we also assume that headers can only be in the top three rows of the table. We noticed manually that there are not many multi-tables among DailyMed drug–drug interaction tables, so it was safe to make this assumption. The algorithm that performed best was the random forest with 97.3% precision and 87.5% recall. In order to perform a test on a new dataset, with the whole methodology, including post-processing, we randomly selected 50 tables for training and 50 tables for testing that were manually inspected whether the headers are classified and annotated correctly. The results are presented in Table 16. Relationships between cells rely on functional analysis, and so we have not modified our original methodology for structural analysis.

**Table 14** Functional analysis evaluation of the original TableDisentangler methodology on the DailyMed subset

|  | TP | FP | FN | Precission | Recall | *F*-Score |
|---|---|---|---|---|---|---|
| Cell role—header | 61 | 39 | 32 | 0.6100 | 0.6559 | 0.6321 |
| Cell role—stub | 309 | 0 | 0 | 1.0000 | 1.0000 | 1.0000 |
| Cell role—super-row | 49 | 6 | 45 | 0.8909 | 0.5213 | 0.6578 |
| Cell role—data | 675 | 18 | 104 | 0.9740 | 0.8664 | 0.9171 |
| Overall (micro average) | 1094 | 63 | 181 | 0.9455 | 0.8580 | 0.9014 |

**Table 15** Machine learning header detection using various algorithms and 10-fold cross-validation on the created dataset

| Algorithm | Precision | Recall | $F$-score |
|---|---|---|---|
| Naive Bayes | 0.588 | 0.936 | 0.722 |
| Bayesian Networks | 0.559 | 0.964 | 0.708 |
| SVM with SMO | 0.985 | 0.821 | 0.896 |
| C4.5 decision tree | 0.944 | 0.307 | 0.463 |
| Random forests | 0.973 | 0.875 | 0.922 |

**Table 16** Machine learning header detection evaluation for the Daily-Med subset

| Dataset | TP | FP | FN | Precision | Recall | $F$-score |
|---|---|---|---|---|---|---|
| Training data | 288 | 8 | 41 | 0.973 | 0.875 | 0.922 |
| Testing data | 176 | 59 | 26 | 0.749 | 0.871 | 0.805 |

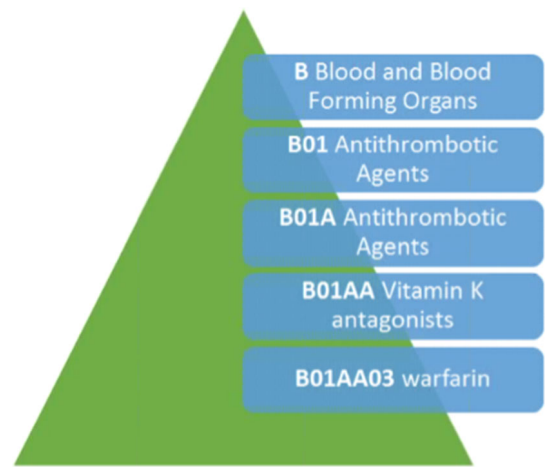

**Fig. 11** ATC coding system

As our training data contained only 823 instances, the algorithm did not manage to learn all the possible table header cues. The training data also contained similar entries, while tables selected for testing contained more diverse cues. In short, our training data were not large enough to learn possible cues and achieve performance closer to the training data. However, the results are significantly better than without using machine learning.

### 5.3 Pragmatic analysis

Tables containing potential drug–drug interactions are only in the section describing drug interactions. This section is labelled with LOINC (Logical Observation Identifiers Names and Codes) code 34073-7, and therefore, pragmatic analysis is trivial using a single rule (whether LOINC code of the section in which table appears is 34073-7).

### 5.4 Table annotation

We annotated cell content using the Unified Medical Language System's (UMLS) and MetaMap program to identify named entities within the table cells [3,5]. The annotation method stored the MetaMap annotations as Concept Unique Identifiers (CUIs) linked to data from specific table cells. The UMLS semantic network provides a semantic type for each CUI, such as Pharmacologic Substance, Clinical Attribute or Therapeutic or Preventative Procedure. These annotations can be further linked to information from UMLS through CUI, such as ATC (The Anatomical Therapeutic Chemical). Using the ATC codes, we can determine on which organ or system a drug's active ingredient is acting and whether the cell is describing a single drug or drug group. Drug codes contain seven characters, while drug codes for drug groups or systems on which a drug is acting contain fewer characters. An example of the coding can be seen in Fig. 11. It is important for users of the drug–drug interaction database to know whether the drug is interacting with the whole drug group or just a single drug ingredient.

### 5.5 Cell selection and syntactic analysis

Once the tables were annotated, we proceeded with crafting rules for extracting drug–drug interaction. We extracted the drug that the drug label described. This was performed without looking at the table as the document contained XML tag that name the drug SPL refers to. As previously discussed, we only looked for tables presented in the section labelled with LOINC 34073-7. In this case study, we are extracting drugs that interact with the drug the label is about. Therefore, we are dealing with a categorical variable. (There is a closed set of possible drugs.) The lexical white list for headers contained words "drug", "coadministered" or "co-administered". The header cell should not contain cues like "effect", "dose", "exposure" or "recommendation" (the lexical blacklist). We selected the column defined by the mentioned keywords. Our method extracts cells below the header in the given column unless the column is spanning, is a super-row or the cell is empty.

Further, the extracted information can be syntactically and semantically analysed in order to obtain one-on-one drug interactions. Often, tables present multiple drugs in one data cell. Authors group cells by drug groups and present multiple drugs from the same group in one cell (see example in Fig. 12). Our extraction methodology extracts the content of the cell as one interaction entry (as it is in the table). However, in case one wants to obtain the pair of drugs that are interacting, further analysis is necessary. UMLS and ATC annotations provide valuable help in obtaining pairs and recognizing drug groups and individual drugs. However, not all drugs and/or drug groups can be annotated. Therefore, the content must be appropriately split. The content that mixes drug/ingredient names with text (for example about dosage) can be challenging to parse and find the interacting drug. The task involves drug named entity recognition and is beyond the scope of this project.

| Table 4 Drugs Tested in *In Vitro* Binding or *In Vivo* Drug Interaction Testing or With Post-Marketing Reports | |
|---|---|
| Drugs with a known interaction with colesevelam | Cyclosporine[c], glyburide[a], levothyroxine[a], and oral contraceptives containing ethinyl estradiol and norethindrone[a] |
| Drugs with postmarketing reports consistent with potential drug–drug interactions when coadministered with WELCHOL | phenytoin[a], warfarin[b] |
| Drugs that do not interact with colesevelam based on **in vitro** or **in vivo** testing | cephalexin, ciprofloxacin, digoxin, warfarin[b] fenofibrate, lovastatin, metformin, metoprolol, pioglitazone, quinidine, repaglinide, valproic acid, verapamil |

**Fig. 12** Example of a table presenting multiple interacting drugs per cell (SetID: b9df447c-b65b-45b9-873a-07a2ab6e2d1f)

**Table 17** Evaluation of potential drug–drug interaction pairs from tables in DailyMed

| Dataset | TP | FP | FN | Precision | Recall | $F$-score |
|---|---|---|---|---|---|---|
| Training data | 514 | 16 | 128 | 0.970 | 0.819 | 0.888 |
| Testing data | 428 | 45 | 122 | 0.904 | 0.778 | 0.836 |

We used 50 randomly selected tables for rule development and an additional 50 tables for evaluation. The evaluation results are presented in Table 17. Our extraction template contained drugs, which the drug label described, interacting drugs and metadata about tables and articles from which data were extracted. If interacting fields contained multiple drugs or drug classes, we assumed correct extraction (true positive). If the algorithm extracted a cell that did not contain interacting drugs or drug classes, we counted it as false positive. If a cell containing interacting drugs or drug classes is missed by the algorithm, it is counted as a false negative.

With an $F1$-score of 0.877 for the training data and 0.836 for the test data, the results are satisfactory for a drug–drug information extraction task.

However, these scores can be improved with further iterations. In both cases, precision is high and there are not too many false positives. The false positives occurred by collecting rows that described drugs in cells below (usually super-row). The false negatives were mainly caused by changes to table structure in which a new header was presented in a row that overrode the initial header. For example, the table may present drugs in the first column, while the effect of the interaction is in the second column. However, in the middle of the table, a new header presents drugs that increase the effect of some substance in the first row and drugs that decrease the effect of the same substance in the second column. This is often changed back to the initial table structure by adding a super-row that groups drugs by target organ or disease. This way of presenting information is used infrequently.

## 5.6 Remarks about the generalizability of framework

The presented case study shows that the approach is generalizable, even to the datasets having a number of unique challenges, with minimal changes. In the following list, we present which steps of the method are generalizable and which need certain modifications:

– *Document and table reading* requires modifications for a given data format. There are defined data structures that have to be populated from the original document. Once the data structures are populated, the methodology requires little modification.
– *Table detection* remains the same across majority of XML documents (finding *table* tags). However, other formats, such as PDF or ASCII text document, may require more complex table detection methodology.
– *Functional analysis* remains the same for the majority of documents that contain emphasis features that would clearly distinguish headers, stubs, super-rows and data cells. In documents with tables not containing enough emphasis cues (different font style, breaking lines, etc.), it may be necessary to introduce some lexical classification of cells, as we did in DailyMed case study.
– *Structural analysis* does not need any modification and depends on the output of functional analysis
– *Pragmatic analysis* can be performed either by rules or by utilizing machine learning classification. It depends on the task; therefore, new rules or new classification models may be necessary for each task.
– *Table annotation* in the biomedical domain, it is standard to use UMLS annotation, and for the most of the tasks, it would be helpful. However, in other domains may be utilized other taxonomies, vocabularies or ontologies.
– *Cell selection* the framework for creating rules remains the same, involving white list and blacklist. However, rules will be different for different tasks.

- *Syntactic processing* many of the data presentation patterns, especially for numeric values, can be reused for many tasks and domains. However, for certain tasks, a new set of syntactic rules have to be crafted. The framework allows easy creation of these rules using regular expressions and assignment of semantics to the extracted value groups.

The theoretical framework is generalizable to different domains, tasks and document formats. Even though the framework is developed with XML documents in mind, it can be extended beyond that format by creating appropriate document readers that can populate table data structures in TableDisentangler. Also, the framework can work with the majority of table layouts, that may include both vertically and horizontally spanning cells. The remaining challenge of creating extraction lexical, semantic and syntactic rule for each task remains.

## 6 Software, datasets and availability

The framework containing the elaborated steps (table detection, functional analysis, structural analysis, pragmatic analysis, table annotation, cell selection and syntactic analysis) is implemented in two tools that have to be executed sequentially and are available open source. Table detection, functional and structural analysis have been previously described in [30] and implemented in TableDisentangler[8] tool. Pragmatic classification and semantic table annotations are also features of TableDisentangler [29].

For defining a variable of interest, defining extraction lexical, semantic and syntactic rules, we developed a wizard-like tool called TableInOut.[9] In the TableInOut, the user is able to define the name of the variable, pragmatic type, lexical and semantic rules with the functional areas where they should be searched for. Also, the user can either create or reuse a set of syntactic rules for extracting atomic values from tables. Using this tool, we were able to reproduce the results presented in Tables 8, 9 and 10.

## 7 Conclusion

Similar to the natural language, that needs to be analysed from lexical, syntactic and semantic perspective [2], tables as well need multilayered analysis. Analysis of table, its content and meaning sometimes require all the steps needed for natural language processing, since the content of the cells is presented in natural language. However, table processing have additional processing layers, that include physical (detection of the table boundaries), functional (recognizing functions of cells and areas within the table), structural (recognizing structure and relationships between the cells), pragmatic (purpose of the table), syntactic (organisation of words, numbers and symbols in cells) and semantic (meaning of presented values) perspectives. Language that is used in tables can be chunked (with omitted words, acronyms and abbreviations) and ungrammatical. Analysis of such language can present a challenge, but also in many cases it will reduce the quite complex syntactic analysis of lexical cues, simplifying that part of the text processing. In the case of the tables that present whole sentences in natural language, lexical processing of cell content cannot be simplified.

Information in tables can be categorized into two broad categories (textual and numeric) or five more narrow ones (single numeric, cumulative statistic values, categorized numerical, textual categorical and free text). Since each of the categories has its specifics, information extraction methodology slightly differs for each of them. They use different annotations, value patterns or have different features (e.g. numerical values may have units of measure, while categorical and textual would not have it). We defined information that person developing information extraction rules for a certain variable in tables need to know, such as binding to the semantic resource, functional, lexical, syntactic cues, possible and default units of measure and pragmatic type. Once this information is known, one can craft rules and iteratively improve them. We tried to automate as many steps as it was possible, so person creating rules for new variables does not need to interact with these parts. Table detection, functional and structural analysis are usually generic, and there is no need for new rules in these processes. However, in cases of some datasets emphasis features of the table are omitted and it is challenging to distinguish functional areas. In these cases, machine learning based on content can help; however, this approach will introduce domain dependence. Semantic tagging can be performed by many tools, vocabularies or ontologies, which require some effort. Still there exist two layers for which user need to define rules—lexical and syntactic.

While developing our methodology for information extraction from tables, we tried to examine in which steps of the methodology machine learning can help compared to a rule-based approach. Several steps were modelled as classification tasks. Machine learning was able to help in pragmatic classification, allowing us to easily create a well-performing model. The performance of the pragmatic classification was dependant on how specific pragmatic classes were (more specific class—better classifier). Due to unbalanced data, it was not easy to create a well-performing classifier for recognizing cell of interest in the information extraction task. Machine

---

learning-based approach was more labour-intensive and performed the same or worse than rule-based approach. Also, due to unbalanced data and the size of the training data, it was necessary to perform additional processing over the data (such as cost-sensitive classification).

Our approach present state-of-the-art in table information extraction from XML documents, without any restriction on the structure of tables. The approach is also generalizable over different domains (we tested on clinical trial and drug label) and over different datasets. Even though some of the previous approaches reported slightly better performance [11,46], they were limited to standardized tables with pre-defined table's structure. The first three steps of our approach are domain- and task-independent. In certain challenging datasets, domain independence may be traded for detection accuracy in the case when table do not emphasize navigational areas using machine learning. Semantic tagging and pragmatic processing are domain dependent, but task independent (semantic tags and pragmatic classes of tables can be used in information extraction, information retrieval and other tasks), while the information extraction rules are domain and task dependent.

# 8 Future work

For the future work beyond this, we propose the following:

1. *Generalization for other document formats* The major limitation of this work is that the presented methodology only supports documents in an XML format. A large amount of the scientific literature is published in PDF, and other document formats are also used frequently. Our methodology provides general heuristic guidelines for functional analysis. However, in order to implement this approach, it is necessary to apply optical character (and object, including lines) recognition and other techniques to transform and disentangle tables from visual representation to the appropriate representation for computational handling. There are a number of tools for converting, for example, PDF documents into XML, such as pdf2xml, pdftohtml, pdfextract, SectLabel, PDFX and easyPDF SDK [8]. These tools might be used more or less successfully in a preparatory step for the PDF format. Once tables are transformed into the proposed data model and stored in a database, it is possible to apply tools developed in this thesis for information extraction.

2. *Evaluate effects of assisted curation* In this work, we have not tested the effects on the speed and accuracy of machine-assisted data curation for table mining. The assumption that it will significantly increase curation speed is based on the literature on assisted curation from the text. (Automated or assisted curation can speed up the curation process by more than 70% [1].) It is left for the future to design the user interface and examine the gains of assisted data curation from tables.

3. *Explore table representations for deep learning* In recent years, deep learning and deep neural networks archived successes in many areas, ranging from playing games to natural language processing [22,38]. Several text vector representation models significantly improved the performance of text classification, named entity recognition and information extraction in text [26,36]. These models are able to handle the linguistic context of the words in the text. However, they are not designed to handle visual structures and make predictions based on them. Vector representations that would involve both context and structure of the article element should be explored in the future. The first attempt to do so was performed by [13]; however, this approach is limited to classification of table clusters, with no relation to other tasks where vector representation may be helpful. Also, the performance of information extraction using recurrent neural networks in combination with the mentioned representation model should be further explored in the future.

4. *Examine other text mining tasks* In the past, approaches in information retrieval [17,24], information extraction [11,31] and knowledge discovery [49,52] from tables were presented. Some text mining tasks, such as relation extraction, summarization, question answering, topic segmentation and recognition have not been examined. These research fields lack research activity, especially in the biomedical domain that is rich in tables that provide valuable information important for experiment reproduction, evidence synthesis and future research. This thesis provides a foundation with a table and data model for table analysis that can be utilized for higher layers of table analysis to solve said tasks. This is especially true for PMC data, for which we provided methods that perform the functional and structural analyses. We briefly mentioned topic recognition in our pragmatic analysis, which recognizes the main table topic. However, the extraction task designer assigned possible topics manually. An automatic and generic topic analysis mechanism that can automate pragmatic analysis remains a task for future development. Question answering relies on information extraction and information retrieval but also employs a number of specific normalization techniques. The specifics of table data and the influence of table structure on question answering also remain tasks for future development. Table summarization may help with large and complex table reading. Summarization would aid a reader in determining whether or not information he/she is looking for may be stored in the table as well as present the most important findings to a user without going into the table (e.g. statistically significant results). This task requires a complex semantic analysis of the table data, as well as analysis of the surrounding text.

**Acknowledgements** This research was funded by Engineering and Physical Sciences Research Council (EPSRC) and AstraZeneca plc.